\documentclass{article} 
\usepackage{iclr2027_conference,times}

\usepackage{amsmath,amsfonts,bm}

\def\eqref#1{equation~\ref{#1}}

\def\1{\bm{1}}

\DeclareMathAlphabet{\mathsfit}{\encodingdefault}{\sfdefault}{m}{sl}
\SetMathAlphabet{\mathsfit}{bold}{\encodingdefault}{\sfdefault}{bx}{n}

\usepackage{hyperref}
\usepackage{url}

\usepackage{cleveref}
\usepackage{amsmath}
\usepackage[most]{tcolorbox}
\tcbuselibrary{breakable}
\definecolor{myblue}{RGB}{0, 102, 204}
\definecolor{lightblue}{RGB}{210, 225, 245}

\usepackage{wrapfig}      
\usepackage{booktabs}
\usepackage{tabularx}
\usepackage{xcolor}
\usepackage{threeparttable}
\usepackage{adjustbox}

\usepackage{multirow}   
\usepackage[table]{xcolor}

\usepackage{rotating}
\usepackage{float}
\usepackage{wrapfig}

\title{Every Token Needs Its Own Teacher: Multi-Teacher Adaptive On-Policy Self-Distillation}

\author{Yifu Luo $^{1,2,3}$, Chuang Ma $^{4,5}$, Zeyu Chen $^{1,2}$, \AND Enkelejda Kasneci $^{1,2}$, Qi Chen $^{6,7}$, Gjergji Kasneci $^{1,2}$ \\
  $^{1}$TU München \vspace{.1em} \hspace{4pt} $^{2}$Munich Center for Machine Learning (MCML) \vspace{.1em} \hspace{4pt} 
  \\
  $^{3}$relAI – Konrad Zuse School of Excellence in Reliable AI \vspace{.1em} \hspace{4pt} $^{4}$Kyoto University \vspace{.1em} \hspace{4pt} $^{5}$NII LLMC \vspace{.1em} \hspace{4pt}
  \\
  $^{6}$ELLIS Institute Finland \vspace{.1em} \hspace{4pt} $^{7}$Aalto University \vspace{.1em} \hspace{4pt}
}

\iclrfinalcopy 
\begin{document}

\newcommand{\threefindings}[3]{%
\begin{tcolorbox}[
    enhanced,
    colback=white,
    colframe=white,
    leftrule=0.4mm,
    rightrule=0.4mm,
    toprule=0.4mm,
    bottomrule=0.4mm,
    arc=0mm,
    left=0pt,
    right=0pt,
    top=2pt,
    bottom=2pt,
    breakable,
    borderline north={0.4mm}{0pt}{myblue},
    borderline south={0.4mm}{0pt}{myblue}
]
{\textbf{\textcolor{myblue}{Findings$1$:}}} #1 \\
{\textbf{\textcolor{myblue}{Findings$2$:}}} #2 \\
{\textbf{\textcolor{myblue}{Findings$3$:}}} #3
\end{tcolorbox}
}

\maketitle
\begingroup
\renewcommand{\thefootnote}{}
\footnotetext{Correspondence to:
Yifu Luo (\texttt{yifu599@gmail.com}) and
Chuang Ma (\texttt{ma.chuang.52h@st.kyoto-u.ac.jp}).}
\addtocounter{footnote}{-1}
\endgroup

\begin{abstract} 
On-policy Self-distillation (OPSD) has proven effective for post-training large language models by constructing a teacher conditioned on privileged references. However, existing approaches keep the teacher fixed either throughout training by consistently exposing it to the full reference, or within each response by adaptively varying partial reference exposure only across responses. In this paper, we argue that neither of them is sufficient. Instead, every token needs its own teacher, as the teacher preference can vary token by token within the same response. Through a teacher-intervention pilot study, we validate that the teacher preference differs across tokens, and reveal that a preferred teacher can be identified for each token through majority voting and KL-based selection. Motivated by these findings, we propose Multi-teacher Adaptive On-policy Self-Distillation (MTA-OPSD), the first token-level adaptive OPSD method. Our method constructs a multi-teacher pool consisting of candidate teachers, and enables every token to adaptively select its preferred teacher during training. Experiments with Qwen3 models on mathematical reasoning tasks show that MTA-OPSD consistently outperforms existing OPSD approaches with an average gain of up to $+5.9$ over the base model, highlighting the importance of adapting teacher supervision at the token level. The code is available at \url{https://github.com/xingzhejun/MTA-OPSD}.
\end{abstract}

\section{Introduction}

On-policy Self-distillation (OPSD) \citep{zhao2026self, hubotter2026reinforcement, shenfeld2026self,cui2026brief} has emerged as a promising paradigm for post-training large language models (LLMs), showing superiority over other post-training paradigms like Supervised Fine-tuning (SFT) and Reinforcement Learning with Verifiable Rewards (RLVR) (e.g., GRPO \citep{guo2025deepseek}). In OPSD, the same model serves as both student and teacher. The student generates rollout responses, while the teacher provides supervision with access to privileged references, such as reference solutions. Compared with RLVR, OPSD provides dense token-level learning signals beyond sparse outcome rewards. Compared with SFT, it trains on trajectories sampled by the student itself, mitigating the exposure bias \citep{bengio2015scheduled}. 

Despite these advantages, recent studies \citep{kim2026does,kaur2026rethinking,harne2026privileged,armandpour2026unmasking} reveal that OPSD can also suffer from degraded reasoning performance, which stems from the highly informative references exposed to the teacher. It may induce overly sharp teacher distributions, producing supervision that is poorly aligned with the student's learning dynamics (e.g., suppressing uncertainty \citep{kim2026does, zhang2026embarrassingly}). A line of work \citep{tan2026paint,han2026adaptive,yang2026matching} has attempted to address this student-teacher mismatch through adaptive reference exposure. Rather than providing the teacher with the complete reference, it adjusts how much of the reference the teacher receives based on the student's learning dynamics, such as the student-teacher gap. However, these methods adapt at the response level. Although reference exposure may vary across responses, it remains fixed across tokens within each response. In other words, all tokens in a rollout are supervised by the same teacher.

In this paper, we challenge this consensus by arguing:
\par\noindent
\hspace{2em}\textit{The adaptive teacher should operate at the token level, because every token requires its own} \\
\hspace*{2em}\textit{dedicated teacher.}
\par
\begin{wrapfigure}{r}{0.54\textwidth}
    \vspace{-20pt}
    \centering
    \includegraphics[width=\linewidth]{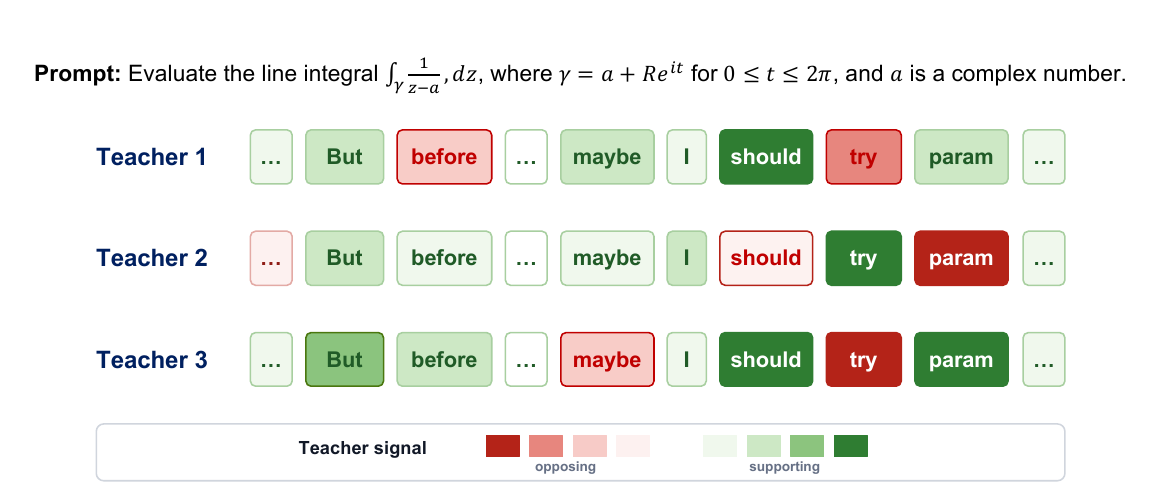}
    \vspace{-10pt}
    \caption{Token-level learning signals induced by different teachers for the same response. Teachers agree at some positions but provide opposing signals at others, motivating token-level teacher selection. Additional details on this visualization are provided in \Cref{a1}.}
    \label{fig1}
    \vspace{-8pt}
\end{wrapfigure}
A teacher that provides useful guidance at one token may be less suitable at another, even within the same response. Rather than treating it with the same teacher, each token in the response merits consideration with its own preferred teacher. As illustrated in \Cref{fig1}, when different teachers are applied to the same response, they can provide similar learning signals for some tokens but disagree on the direction of supervision for others. This suggests that not only each response, but also each token, may have distinct teacher preferences.

\begin{wrapfigure}{r}{0.6\textwidth}
    \vspace{-6pt}
    \centering
    \includegraphics[width=\linewidth]{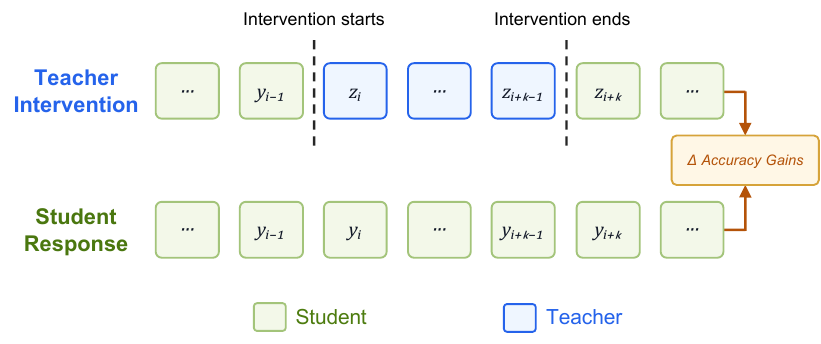}
    \vspace{-12pt}
    \caption{Illustration of the teacher-intervention pilot study.}
    \label{fig2}
    \vspace{-12pt}
\end{wrapfigure}

While some prior work \citep{wang2026beyond,hao2026rethinking,xu2026tip,ke2026respecting,wei2026amr,wang2026trace,li2026unifying,han2026distill} has explored token selection for post-training (e.g., focusing solely on high-entropy tokens \citep{wang2026beyond}), how to validate our claim is challenging, and how to adaptively identify the preferred teacher for each token remains unexplored. To answer these, we conduct a teacher-intervention pilot study (\Cref{sec4}) to quantify the token-level teacher preference. As shown in \Cref{fig2}, we preserve a student-generated prefix, allow a candidate teacher to generate a short continuation at a selected token position, and then return control to the student to complete the response. The accuracy gain from this ``student-teacher-student" relay over pure student generation serves as a proxy for the teacher preference at that token position. A large gain implies a compatible teacher for that position, while a small gain suggests the opposite. Interestingly, the results are simple, yet clear and compelling. First, teacher preferences differ across tokens, even within the same response. Second, at the token level, majority voting yields a subset of reliable teachers, and the preferred teacher always exhibits a large student-teacher KL distance. The first finding validates our claim, and the second finding motivates a practical rule for token-level teacher selection.

Based on these findings, we propose Multi-teacher Adaptive On-policy Self-Distillation (MTA-OPSD), the first token-level adaptive OPSD method. As shown in \Cref{fig3}, MTA-OPSD constructs a multi-teacher pool consisting of candidate teachers, by varying the truncation strategy and exposure ratio of the privileged reference. During training, rather than using a fixed teacher for all tokens in the same response, each token adaptively selects its own teacher from the multi-teacher pool based on its preference, through majority voting followed by KL-based selection. Our token-level adaptive OPSD design is simple yet effective, precisely addressing the student-teacher mismatch. By assigning each token its preferred teacher, MTA-OPSD provides a more balanced learning signal rather than an overly sharp, rigid teacher distribution from a single teacher. We evaluate MTA-OPSD on mathematical reasoning benchmarks using models from the Qwen3 family \citep{yang2025qwen3}, and it consistently outperforms existing OPSD approaches with an average gain of up to $+5.9$ over the base model. 

Our contributions are summarized as follows:

\begin{itemize}
    \item We quantify the teacher preference across tokens via a teacher-intervention pilot study, validating that preferred teachers can differ from token to token within the same response. We further reveal that the token-level preferred teacher can be identified through majority voting and KL-based selection. 
    \item We propose MTA-OPSD, the first token-level adaptive OPSD method. By constructing a multi-teacher pool as the supervision source, our approach enables each token to adaptively select its preferred teacher under the guidance of majority voting and KL divergence, effectively addressing the student-teacher mismatch.
    \item We conduct extensive experiments across various models on mathematical reasoning tasks, demonstrating that our approach achieves superior reasoning performance compared to existing OPSD methods. Furthermore, we empirically analyze comprehensive ablation studies, paving the way for future advances in this field.  
\end{itemize}

\section{Related Work}

\subsection{On-policy Distillation and On-Policy Self-distillation}

Knowledge distillation \citep{hinton2015distilling} transfers knowledge from a large teacher model to a student model by training on the teacher's soft output distributions. Early approaches \citep{kim2016sequence, jiao2020tinybert, wang2020minilm} leveraged it to sequence-level distillation, establishing the dominant off-policy distillation approaches. \citep{gu2024minillm, agarwal2024policy, lu2025onpolicydistillation, yang2026learning} extended it to On-policy Distillation (OPD), addressing the exposure bias \citep{bengio2015scheduled} mismatch by shifting the training distribution to the student's own generations.

Despite the rapid progress of OPD, the paradigm relies on a separate, typically stronger teacher model, which may be unavailable or expensive to maintain. To address this challenge, OPSD \citep{zhao2026self, hubotter2026reinforcement, shenfeld2026self,luo2026learning,yang2026self} removes this requirement by instantiating the teacher and student from the same model. Under this setting, the teacher and student share the same model, but the teacher is granted access to a privileged reference unavailable to the student. 

However, recent studies \citep{kim2026does,wang2026denser,kaur2026rethinking,harne2026privileged,shen2026purified} reveal a persistent mismatch in OPSD. In particular, a highly informative reference can induce an overly sharp teacher distribution and create a mismatch between the teacher's supervision and the student's learning dynamics. These observations motivate the line of research for adaptive OPSD methods.

\subsection{Adaptive On-Policy Self-distillation}

Adaptive OPSD seeks to align the teacher supervision with the student's learning dynamics. Rather than exposing the complete privileged reference, existing methods \citep{tan2026paint,han2026adaptive,yang2026matching} control its content or exposure ratio using mechanisms such as learned reference-exposure schedules. By varying the amount of privileged references available to the teacher, these methods aim to avoid overly sharp supervision while retaining the benefits of dense learning signals.

However, these methods adapt at the response level, where all tokens in the response share the same teacher and the same privileged reference. In contrast, we identify that each token needs its own teacher as tokens have distinct teacher preferences (\Cref{sec4}), and further propose our token-level multi-teacher adaptive OPSD method.

\section{Preliminaries}

OPSD is a special case of OPD. OPD transfers knowledge from a stronger teacher model $p_T$ to a weaker student model $p_\theta$ by enforcing dense supervision over trajectories sampled by the student. Given a prompt $x$, the student samples a response $y = \{y^1, y^2, \cdots, y^L\}$, where $L$ denotes the generation length and $y^i$ refers to the $i$-th token in the response. The learning objective is to minimize the token-level KL between the teacher's and the student's next-token distributions:
\begin{equation}
\begin{aligned}
    \mathcal{L}_\text{OPD} (\theta) &= \mathbb{E}_x \left[ \sum_{i=1}^L \mathcal{D}_\text{KL} \left(p_\theta \left( \cdot \mid y^{<i}, x  \right) \mid \mid p_T \left( \cdot \mid y^{<i}, x \right) \right) \right] \\
    &= \mathbb{E}_x \left[ \sum_{i=1}^L \mathcal{D}_\text{KL} \left(p_\theta \left( \cdot \mid c_i  \right) \mid \mid p_T \left( \cdot \mid c_i \right) \right) \right]
\end{aligned}
\label{eq1}
\end{equation}
where $p(\cdot \mid y^{<i}, x)$ denotes the distribution over the next token $y^i$, and we let $c_i$ refer to $(y^{<i}, x)$ for simplicity. While forward KL and reverse KL are most used, other distribution divergence measures like generalized Jensen-Shannon divergence \citep{agarwal2024policy} can also be employed.

Recent advances have extended OPD to OPSD, where the student and teacher are instantiated from the same model $p_\theta$. The difference lies entirely in their conditioning contexts. Privileged reference $r$, such as reference solutions \citep{zhao2026self} or environment feedback \citep{hubotter2026reinforcement}, is appended to the original prompt to construct a teacher-specific prompt. Thus, the teacher distribution is:
\begin{equation}
    p_T (\cdot \mid c_i) = p_\theta \left( \cdot \mid c_i, r \right).
\label{eq2}
\end{equation}
Consequently, the learning objective in \Cref{eq1} adapts into the following:
\begin{equation}
    \mathcal{L}_\text{OPSD} (\theta) = \mathbb{E}_x \left[ \sum_{i=1}^L \mathcal{D}_\text{KL} \left(p_\theta \left( \cdot \mid c_i  \right) \mid \mid p_\theta \left( \cdot \mid c_i,r \right) \right) \right].
\label{eq3}
\end{equation}
In this setting, both the teacher and student share the same model but differ only in the conditioning contexts, and the response is solely generated from the student. 

\section{Pilot Study: Token Matters} \label{sec4}

Before diving into a detailed design, a critical question must first be answered: 
\par\noindent
\hspace{1em}\textit{Given $c_i = (y^{<i},x)$, how can we determine if one teacher $p_T(c_i)$ is more suitable than another?}
\par
A direct approach would be to  train on every token separately with specific teachers, covering enough token-teacher combinations. However, it is obviously not practical. 

We instead answer this question from another view. Motivated by \citep{xu2025speculative,xu2026pass}, we estimate teacher suitability through teacher intervention. Specifically, given a prompt $x$ and a student's response $y=\{y^1, y^2, \cdots, y^L\}$, we select a token position $1 < i < L$ and preserve the student-generated prefix before that position. Starting from position $i$, a candidate teacher takes over the generation process and produces a short subsequence:
\begin{equation}
\begin{aligned}
    z_{1:i-1} &= y_{1:i-1} = p_\theta (x)     \\
    z_{t:t+k} &= p_T (y^{<i},x) = p_\theta (c_i,r),
\end{aligned}
\label{eq4}
\end{equation}
where $k$ denotes the teacher-intervention length. After this intervention, control is returned to the student, which completes the remainder of the response:
\begin{equation}
    z_{i+k+1:L} = p_\theta(z^{<i+k+1}, x).
\label{eq5}
\end{equation}
Thus, we obtain a ``student-teacher-student" relay response $z=\{z_1,z_2,\cdots,z_L\}$. We compare its accuracy with the pure student response, which serves as a proxy for the candidate teacher preference at the selected token position. A larger gain indicates a more suitable teacher, whereas a smaller gain indicates the opposite.

The main purpose of this pilot study is that, by varying both the intervention token position and the candidate teacher, this study separates variation along the token and teacher axes. This allows us to examine whether teacher preferences change across token positions and to identify their patterns. Interestingly, the pattern is simple but clear. We next describe the experimental setup and then present our findings, as highlighted in \Cref{sec4.2}.

\subsection{Setup}

Our study varies two principal factors: tokens at which the intervention begins, and teachers which are used for the intervention. 

For the token axis, we consider both regular-spaced and statistic-extreme token positions. For the former, we select tokens located at $25\%$, $50\%$, and $75\%$ of the original student rollout. For the latter, we select tokens associated with the highest and lowest student entropy, and the largest and smallest student-teacher KL distance \footnote{For identifying these KL-based positions, the teacher is conditioned on the full reference.}. In total, we select $7$ intervention positions from each rollout.

For the teacher axis, we vary both the reference-truncation strategy\footnote{The truncation strategy determines which part of the reference is retained. For example, prefix truncation with an exposure ratio of $0.5$ retains the first half of the reference. The final answer is retained under every truncation setting to ensure a fair comparison.} and the reference-exposure ratio. We consider prefix, suffix, and random-span for truncation strategy, each with exposure ratios of $0.5$ and $0.75$, together with a full-reference teacher. This produces 7 candidate teachers. 

We evaluate intervention lengths across $k=64$, $128$, and $256$. We sample $64$ prompts of at least medium difficulty\footnote{For each prompt, we first sample $4$ student responses and exclude prompts for which all four responses are correct.} from DeepMath-103k \citep{he2026deepmath}. We conduct the study using Qwen3-1.7B and Qwen3-4B as student models. To reduce sampling variance, for every experimental configuration, we sample $2$ teacher continuations and then $2$ student relays following each continuation. In total, $64*7*7*3*2*2*2=75264$ generations are evaluated. 

\subsection{Results and Findings} \label{sec4.2}

Interestingly, the results show clear and meaningful findings.
\threefindings{Teacher preference varies across tokens.}{Majority voting provides a subset of reliable teachers.}{The preferred teacher always exhibits a large KL distance from the student.}
\begin{wraptable}{r}{0.6\textwidth}
  \vspace{-12pt}
  \centering
  \setlength{\tabcolsep}{3mm}
  \caption{Overlap of preferred teachers between tokens within the same rollout.}
  \vspace{0.15cm}
  \label{tab1}
    \begin{threeparttable}[b]
      \begin{tabular}{cccc}
        \toprule
        \textbf{Token Position} & \textbf{0.25} & \textbf{0.50} & \textbf{0.75}\\
        \midrule
        \textbf{0.25} & - & 48.44\% & 50.00\% \\
        \textbf{0.50} & 48.44\% & - & 51.56\% \\
        \textbf{0.75} & 50.00\% & 51.56\% & - \\
        \bottomrule
      \end{tabular}
    \end{threeparttable}
    \vspace{-2pt}
\end{wraptable}

\textbf{Teacher preference varies by tokens. } We define the preferred teacher at a token position as the candidate that produces the largest intervention gain. To determine whether this preference changes within a response, we measure how often the preferred teacher is shared by different regular-spaced tokens in the same rollout. As shown in \Cref{tab1}, for Qwen3-1.7B with $k=256$, the overlap is only approximately 50\%, which indicates that the teacher preference differs by tokens. These results directly support our claim: every token needs its own teacher.

\textbf{Majority voting identifies reliable teachers. } By comparing the teacher distribution with the student one over a student-generated token $y^i$, we can obtain the advantage signal:
\begin{equation}
    a_{T}(y^i \mid c_i) = \text{log} \left(p_{T} \left(y^i \mid c_i \right) \right) - \text{log} \left( p_\theta \left(y^i \mid c_i \right) \right)= \text{log} \frac{p_T (y^i \mid c_i)}{p_\theta (y^i \mid c_i)}.
\label{eq6}
\end{equation}
\begin{wraptable}{r}{0.60\textwidth}
\vspace{-0.6cm}
\centering
\setlength{\tabcolsep}{2.5mm}
\caption{Fraction of preferred teachers contained in the majority subset.}
\label{tab2}
\vspace{0.1cm}
\begin{tabular}{lcc}
    \toprule
    \textbf{Model \textbackslash Token Position}
    & \textbf{Regular}
    & \textbf{Extreme} \\
    \midrule
    \textbf{Qwen3-1.7B} & 77.94\% & 88.09\% \\
    \textbf{Qwen3-4B}   & 87.11\% & 90.68\% \\
    \bottomrule
\end{tabular}

\vspace{0.15cm}

\caption{Average gains of teachers in the majority and minority subsets.}
\label{tab3}
\vspace{0.1cm}
\begin{tabular}{lccc}
    \toprule
    \textbf{Group}
    & \textbf{Majority}
    & \textbf{Minority}
    & \textbf{Overall} \\
    \midrule
    \textbf{Gain} & +3.34\% & +1.88\% & +2.34\% \\
    \bottomrule
\end{tabular}
\vspace{-0.6cm}
\end{wraptable}
$a_T(\cdot \mid c_i)$ indicates the teacher's relative support for the student's choice $y^i$. Specifically, If $a_T(\cdot \mid c_i) > 0$, the teacher favors the sampled token; 
otherwise, it indicates that the student's choice is relatively opposed by the teacher. 
Thus, at each token position, we can divide the candidate teachers into two groups based on this sign, and obtain the majority subset.

We then examine whether the teacher producing the largest gain belongs to the majority subset. As shown in \Cref{tab2}, in around $80\%-90\%$ of cases it falls within this subset. We further compare the average gain of teachers in both the majority and minority subsets. For Qwen3-4B with $k=256$, as shown in \Cref{tab3}, the majority subset always achieves larger gains per teacher. These results indicate that majority voting provides a reliable first-stage filter for teacher selection.
\begin{wraptable}{r}{0.6\textwidth}
  \vspace{-0.4cm}
  \centering
  \setlength{\tabcolsep}{3mm}
  \caption{Average gains from the farthest and nearest teachers.}
  \vspace{0.1cm}
  \label{tab4}
    \begin{threeparttable}[b]
      \begin{tabular}{lccc}
        \toprule
        \textbf{Model \textbackslash Teacher} & \textbf{Farthest} & \textbf{Nearest} & \textbf{$\Delta$} \\
        \midrule
        \textbf{Qwen3-1.7B} & +2.42\% & +0.59\% & +1.83\% \\
        \textbf{Qwen3-4B} & +2.78\% & +1.21\% & +1.57\% \\
        \bottomrule
      \end{tabular}
    \end{threeparttable}
\end{wraptable}

\textbf{Large KL distinguishes the preferred teacher. } We also found an interesting correlation between the teacher preference and the student-teacher KL distance. For each token, we select the teachers with the largest and smallest student-teacher KL distance from the majority subset and compare their gains. As shown in \Cref{tab4}, in regular token positions, the farthest teacher consistently produces greater average gains. This observation suggests that, student-teacher KL distance provides a useful criterion for teacher preference, for each token.

\section{Method}

\begin{figure}[t] 
    \centering 
    \includegraphics[width=1.0\textwidth]{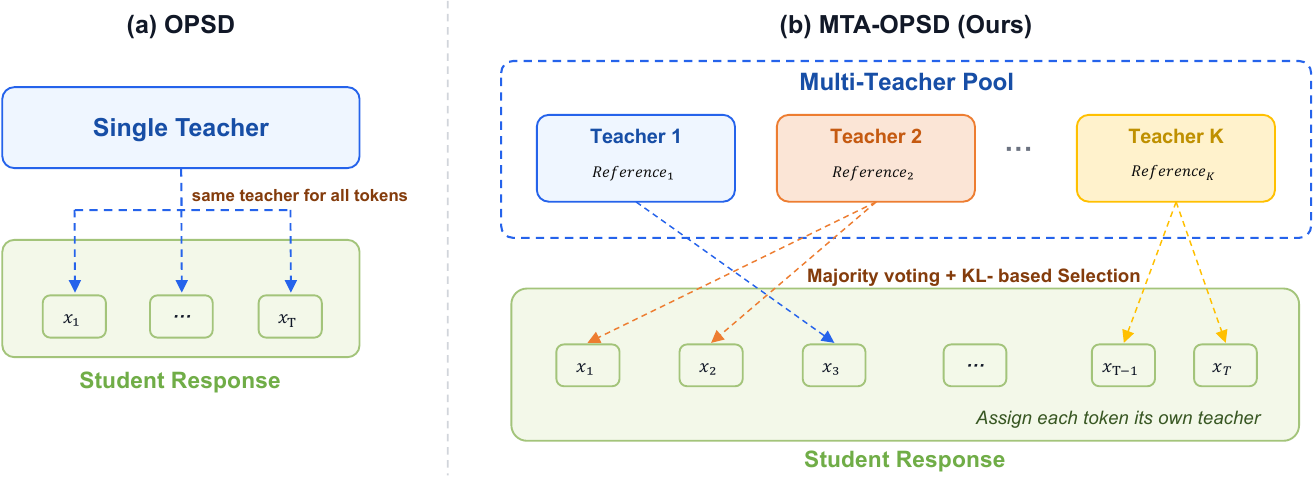}
    \vspace{-12pt}
    \caption{The framework of our approach, MTA-OPSD. With the multi-teacher pool, it adaptively assigns each token its preferred teacher, by majority voting followed by KL-based selection.}
    \vspace{-0.2cm}
    \label{fig3}
\end{figure}

Based on the findings in \Cref{sec4.2}, we propose Multi-Teacher Adaptive OPSD (MTA-OPSD), a simple yet effective framework for token-level adaptive OPSD. As illustrated in \Cref{fig3}, rather than using a single teacher, MTA-OPSD constructs a multi-teacher pool consisting of candidate teachers conditioned on different variants of the privileged reference. Through majority voting followed by KL-based selection, it assigns each token its preferred teacher. This ensures that different tokens within the same response receive supervision from their respective teachers.

\subsection{Multi-teacher Pool}

To support token-level adaptive teacher selection, we construct a multi-teacher pool $M$, consisting of all candidate teachers available for each token:
\begin{equation}
    M = \{p_{T_1},p_{T_2},\cdots,p_{T_K}\},
\label{eq7}
\end{equation}
where $K$ is the number of candidate teachers and $p_{T_k}$ denotes the $k$-th teacher. All teachers share the same model $p_\theta$ with the student. They differ from each other only in the privileged references on which they are conditioned:
\begin{equation}
    R = \{r_{1},r_{2},\cdots,r_{K}\},
\label{eq8}
\end{equation}
which leads to the next-token distribution of each teacher as:
\begin{equation}
    p_{T_k} = p_\theta( \cdot \mid x,r_k).
\label{eq9}
\end{equation}
Each $r_k$ differs from each other by its truncation way and exposure ratio of the full reference. An example of our multi-teacher pool is provided in \Cref{a5}.

While the multi-teacher pool supplies a diverse set of supervision signals, it does not determine which teacher should supervise each token. Guided by the pilot study in \Cref{sec4}, MTA-OPSD performs token-level adaptive teacher assignment in two steps: (1) majority voting identifies a reliable subset of teachers, and (2) KL-based selection chooses the preferred teacher from that subset.

\subsection{Majority Voting} \label{sec5.2}

As shown in \Cref{sec4.2}, majority voting provides a reliable teacher subset by identifying the direction of the advantage signals. We therefore use majority voting as the first step of teacher selection.

Specifically, for a token $y^i$ in the student response, the multi-teacher pool produces the following set of its teacher token distributions:
\begin{equation}
\begin{aligned}
    P_M (c_i) &= \left\{p_{T_1}(\cdot \mid c_i),p_{T_2}(\cdot \mid c_i),\cdots,p_{T_K}(\cdot \mid c_i) \right\}  \\
    &= \left\{p_\theta(\cdot \mid c_i, r_1),p_\theta(\cdot \mid c_i, r_2),\cdots,p_\theta(\cdot \mid c_i, r_K) \right\}.
\end{aligned}
\label{eq10}
\end{equation}
For each teacher token distribution in $P_M$, we compare it with the student token distribution $p_\theta (\cdot \mid y^{<i}, x)$ over $y^i$, to obtain the following set:
\begin{equation}
    A_M (c_i) = \{a_{T_1}(y^i \mid c_i),a_{T_2}(y^i \mid c_i),\cdots,a_{T_K}(y^i \mid c_i)\},
\label{eq11}
\end{equation}
where advantage $a_{T_k}(y^i \mid c_i)$ is computed using \Cref{eq6}. The sign of $a_{T_k}(y^i\mid c_i)$ indicates the $k$-th teacher's relative support for the student's choice at $y^i$. We denote it as the supporting binary vote:
\begin{equation}
    v_{T_k}(c_i)
    =
    \operatorname{sgn}\left(a_{T_k}\left(y^i \mid c_i\right)\right)
    =
    \begin{cases}
        +1, & a_{T_k}\left(y^i \mid c_i\right) > 0, \\
        -1, & a_{T_k}\left(y^i \mid c_i\right) < 0.
    \end{cases}
\label{eq12}
\end{equation}
We then determine the majority voting direction at token position $i$:
\begin{equation}
    V_M (c_i) = \operatorname{sgn} \left( \sum_{k=1}^K v_{T_k} \left(c_i \right)  \right),
\label{eq13}
\end{equation}
and the corresponding token-level majority teacher subset\footnote{If a tie occurs between the majority and minority votes, we then compare which side has the larger absolute value. Additional details are provided in \Cref{a2}.}:
\begin{equation}
    M_{\mathrm{maj}} (c_i) = \left\{ p_{T_k} \in M \mid v_{T_k} \left( c_i \right)= V_M \left( c_i \right) \right\}.
\label{eq14}
\end{equation}
Note that this procedure is token-level, performed independently at every token position.

\subsection{KL-based Selection}

Majority voting filters out teachers whose supervision direction disagrees with most candidates, but it does not distinguish among teachers within the majority subset. As shown in \Cref{sec4.2}, the preferred teacher always has large KL distance from the student. We therefore use KL distance as the second-stage selection criterion.

Specifically, for token $y^i$ in the student response and its majority teacher subset $M_{\mathrm{maj}}(c_i)$, the token-level preferred teacher is defined as:
\begin{equation}
    p_{T}^* (c_i) = p_\theta (\cdot \mid c_i, r^*) = \arg\max_{p_{T_k} \in M_{\mathrm{maj}}(c_i)} \mathcal{D}_\text{KL} \left(p_\theta \left( \cdot \mid c_i  \right) \mid\mid p_{T_k} \left( \cdot \mid c_i \right) \right).
\label{eq15}
\end{equation}
Note that since the preferred teacher is determined separately for every token, it can vary across tokens within the same response. Consequently, substituting this token-level teacher into the learning objective in \Cref{eq3} gives:
\begin{equation}
\begin{aligned}
    \mathcal{L}_\text{MTA-OPSD} (\theta) &= \mathbb{E}_x \left[ \sum_{i=1}^L \mathcal{D}_\text{KL} \left(p_\theta \left( \cdot \mid c_i  \right) \mid\mid p_T^* \left( \cdot \mid c_i \right) \right) \right] \\
    &= \mathbb{E}_x \left[ \sum_{i=1}^L \mathcal{D}_\text{KL} \left(p_\theta \left( \cdot \mid c_i  \right) \mid\mid p_\theta \left( \cdot \mid c_i,r^* \right) \right) \right],
\end{aligned}
\label{eq16}
\end{equation}
Thus, MTA-OPSD assigns each token its own teacher.

\subsection{Discussion}

The mechanism of our method can be understood as decoupling the reliability of the supervision direction from the informativeness of the supervision signal. Majority voting first determines whether the teacher pool collectively supports or opposes the student-generated token $y^i$, KL-based selection then identifies the most informative teacher aligned with that direction. This ordering is crucial: KL distance is not used as a standalone measure of teacher quality, but only after directional consensus has filtered out potentially unreliable teachers.

Specifically, when $V_M(c_i)=1$, the majority teachers in $M_{\mathrm{maj}}(c_i)$ assign a higher probability to $y^i$ than the student itself, and selecting the teacher with the largest KL therefore reinforces the student's current choice. Meanwhile, the selected teacher can also provide a substantially different overall probability distribution. Consequently, the supervision can strengthen the sampled token without being limited to simply imitating it, thereby retaining informative probabilities for other plausible predictions and potentially reducing overfitting to the student's current trajectory.

When $V_M(c_i)=-1$, the majority teachers assign a lower probability to $y^i$, and the sampled token is collectively regarded as a less preferred choice. In this case, a teacher with a small KL distance may only weakly downweight this token or remain too close to the student distribution. In contrast, the farthest teacher is more likely to redistribute probability mass toward different candidate tokens, providing a stronger corrective signal together with more distinct alternative prediction directions. Such supervision may therefore help the student move away from a potentially erroneous local choice rather than merely making a marginal adjustment to it.

\section{Experiments}

\subsection{Experiment Setup} \label{sec6.1}

Additional experiment setup details are provided in \Cref{a3}.

\textbf{Models and Datasets. } We evaluate models from the Qwen3 family \citep{yang2025qwen3} at three scales: Qwen3-1.7B, Qwen3-4B, and Qwen3-8B, all with thinking mode enabled during both training and validation. We conduct experiments across mathematical reasoning tasks. The training data
is drawn from DeepMath-103K \citep{he2026deepmath}, and we evaluate on AMC23 \citep{amc23}, AIME24 \citep{aime24}, and AIME25 \citep{aime25}. 

\textbf{Baselines. } We compare our method against two fixed-teacher OPSD methods and one adaptive OPSD method: (1) OPSD \citep{zhao2026self}, which performs dense-logit distillation using forward KL with token-wise pointwise KL clipping; (2) SDPO \citep{hubotter2026reinforcement}, which performs dense-logit distillation using Jensen–Shannon divergence; and (3) Paint \citep{tan2026paint}, which adaptively determines the reference-exposure ratio according to the student-teacher gap.

\textbf{Implementation Details. } All baselines and our method use dataset-provided ground-truth reference solutions as the privileged information, and all contexts include both the reasoning trace and the final answer. We use identical training and evaluation settings across all methods, with maximum generation lengths of 4096 tokens during training and 38912 tokens during validation. The multi-teacher pool contains seven teacher configurations: \{full reference, prefix-$0.5$, prefix-$0.75$, suffix-$0.5$, suffix-$0.75$, random-span-$0.5$, random-span-$0.75$\}\footnote{The truncation strategy determines which part of the reference is retained. For example, prefix truncation with an exposure ratio of $0.5$ retains the first half of the reference. The final answer is retained under every truncation setting to ensure a fair comparison.}. We train each method for 300 steps, evaluate checkpoints every 50 steps on the test set, and report its best Avg@12 results.

\subsection{Main Results}

\begin{table}[!t]
\vspace{-0.5cm}
\centering
\caption{Mathematical reasoning performance across model scales. MTA-OPSD achieves the best average performance at all three model scales.}
\vspace{0.1cm}
\begin{threeparttable}
\begin{adjustbox}{max width=\textwidth}
\begin{tabular}{llcccc}
\toprule
\textbf{Model} & \textbf{Method} & \textbf{AMC23} & \textbf{AIME24} & \textbf{AIME25} & \textbf{Average} \\
\midrule

\multirow{5}{*}{Qwen3-1.7B}
& Base Model & 74.1 & 45.6 & 33.6 & 51.1 \\
& OPSD & 75.3 & 49.2 & 37.5 & 54.0 \\
& SDPO & 73.5 & 43.3 & 31.4 & 49.4 \\
& Paint & 76.0 & 47.5 & 38.6 & 54.0 \\
& \textbf{MTA-OPSD} & \cellcolor{lightblue} 78.5 & \cellcolor{lightblue} 51.4 & \cellcolor{lightblue} 41.1 & \cellcolor{lightblue} 57.0$_{\textcolor{green!60!black}{\scriptsize +5.9}}$ \\

\midrule

\multirow{5}{*}{Qwen3-4B}
& Base Model & 88.6 & 73.1 & 65.3 & 75.7 \\
& OPSD & \cellcolor{lightblue} 89.1 & \cellcolor{lightblue} 74.4 & 65.6 & 76.4 \\
& SDPO & 87.7 & 71.4 & 61.7 & 73.6 \\
& Paint & \cellcolor{lightblue} 89.1 & 71.7 & 66.1 & 75.6 \\
& \textbf{MTA-OPSD} & 89.0 & \cellcolor{lightblue} 74.4 & \cellcolor{lightblue} 68.1 & \cellcolor{lightblue} 77.2$_{\textcolor{green!60!black}{\scriptsize +1.5}}$ \\

\midrule

\multirow{5}{*}{Qwen3-8B}
& Base Model & 89.1 & 73.1 & 66.9 & 76.4 \\
& OPSD & 89.9 & 75.0 & 67.8 & 77.6 \\
& SDPO & 88.9 & 73.9 & 65.6 & 76.1 \\
& Paint & \cellcolor{lightblue} 90.0 & 76.1 & 68.3 & 78.1 \\
& \textbf{MTA-OPSD} & 89.8 & \cellcolor{lightblue} 77.5 & \cellcolor{lightblue} 70.3 & \cellcolor{lightblue} 79.2$_{\textcolor{green!60!black}{\scriptsize +2.8}}$ \\

\bottomrule
\end{tabular}
\end{adjustbox}
\end{threeparttable}
\label{tab5}
\vspace{-10pt}
\end{table}

\Cref{tab5} presents a comprehensive reasoning performance comparison between existing OPSD methods and our approach. For average performance, MTA-OPSD achieves the highest average performance at every model scale, yielding an average improvement of $+5.9$ for Qwen3-1.7B, $+1.5$ for Qwen3-4B, and $+2.8$ for Qwen3-8B, over the base model. These results outperform existing OPSD methods. For individual benchmarks, MTA-OPSD also achieves or matches the best result on nearly every model-benchmark combination. Note that the improvements are particularly consistent and strong on the more challenging AIME25 benchmark. These results highlight the importance of token-level adaptive teacher assignments. A qualitative comparison is provided in \Cref{a4}.

\subsection{Ablation Studies} \label{sec6.3}

All ablation studies use Qwen3-1.7B as the base model and follow the same experiment settings.
\begin{wraptable}{r}{0.66\textwidth}
    \vspace{-10pt}
    \centering
    \setlength{\tabcolsep}{2mm}
    \small
\caption{Comparison of MTA-OPSD with single-teacher variants.}
\label{tab6}
\begin{adjustbox}{max width=\linewidth}
\begin{tabular}{lcccc}
    \toprule
    \textbf{Teacher}
    & \textbf{AMC23}
    & \textbf{AIME24}
    & \textbf{AIME25}
    & \textbf{Average} \\
    \midrule
    Base Model       & 74.1 & 45.6 & 33.6 & 51.1 \\
    Full (OPSD)      & 75.3 & 49.2 & 37.5 & 54.0 \\
    Prefix-$0.5$     & 76.8 & 50.8 & 40.0 & 55.9 \\
    Prefix-$0.75$    & 78.2 & 49.4 & 41.1 & 56.2 \\
    Suffix-$0.5$     & 75.9 & 48.6 & 34.7 & 53.1 \\
    Suffix-$0.75$    & 75.0 & 50.3 & 40.0 & 55.1 \\
    Random-span-$0.5$  & 76.8 & 49.4 & 36.9 & 54.4 \\
    Random-span-$0.75$ & 76.4 & 50.6 & 35.8 & 54.3 \\
    \textbf{MTA-OPSD}
    & \cellcolor{lightblue} 78.5
    & \cellcolor{lightblue} 51.4
    & \cellcolor{lightblue} 41.1
    & \cellcolor{lightblue} 57.0 \\
    \bottomrule
\end{tabular}
\end{adjustbox}


\caption{Comparison of token-level teacher-assignment strategies using the same multi-teacher pool.}
\label{tab7}
\vspace{0.1cm}
\begin{adjustbox}{max width=\linewidth}
\begin{tabular}{lcccc}
    \toprule
    \textbf{Strategy}
    & \textbf{AMC23}
    & \textbf{AIME24}
    & \textbf{AIME25}
    & \textbf{Average} \\
    \midrule
    Base Model  & 74.1 & 45.6 & 33.6 & 51.1 \\
    Random-pick & 76.1 & 49.2 & 35.5 & 53.6 \\
    Mean-loss   & 75.7 & 48.9 & 36.4 & 53.7 \\
    \textbf{MTA-OPSD}
    & \cellcolor{lightblue} 78.5
    & \cellcolor{lightblue} 51.4
    & \cellcolor{lightblue} 41.1
    & \cellcolor{lightblue} 57.0 \\
    \bottomrule
\end{tabular}
\end{adjustbox}

\caption{Ablation of majority voting and KL-based teacher selection.}
\label{tab8}
\begin{adjustbox}{max width=\linewidth}
\begin{tabular}{llcccc}
    \toprule
    \textbf{Majority Voting}
    & \textbf{KL Selection}
    & \textbf{AMC23}
    & \textbf{AIME24}
    & \textbf{AIME25}
    & \textbf{Average} \\
    \midrule
    No & Farthest
    & 75.8 & 48.9 & 33.9 & 52.9 \\
    \midrule
    \multirow{3}{*}{\textbf{Yes}}
    & \textbf{Farthest}
    & \cellcolor{lightblue} 78.5
    & \cellcolor{lightblue} 51.4
    & \cellcolor{lightblue} 41.1
    & \cellcolor{lightblue} 57.0 \\
    & Random
    & 78.0 & 51.1 & 40.8 & 56.6 \\
    & Nearest
    & 75.7 & 48.6 & 35.6 & 53.3 \\
    \bottomrule
\end{tabular}
\end{adjustbox}
\vspace{-15pt}
\end{wraptable}

\textbf{Effect of the Multi-Teacher Pool. } We first investigate whether the multi-teacher pool provides more effective supervision than any individual teacher. In \Cref{tab6}, we compare our method with training using each teacher configuration from the pool independently. Although the effectiveness of individual teachers varies from each other, MTA-OPSD achieves the highest score on both average and individual benchmarks. These results suggest that the multi-teacher pool offers diverse, sufficient, and complementary learning signals. 


\textbf{Effect of Token-Level Teacher Assignment. } The multi-teacher pool is accompanied by the token-level adaptive teacher assignment. The previous ablation establishes the value of the former, but does not isolate the contribution of the latter. We therefore retain the same multi-teacher pool and compare MTA-OPSD with two alternative assignment strategies: (1) Random teacher selection from the pool for each token, and (2) averaging the distillation losses from all teachers. As shown in \Cref{tab7}, MTA-OPSD consistently outperforms both alternatives on every benchmark. These show that the gains do not arise solely from multiple teachers; selecting a suitable teacher for each token is critical to using the teacher pool effectively.

\textbf{Teacher-Assignment Dynamics. } We also investigate how frequently each teacher is selected throughout training. The assignments are distributed relatively evenly across the teacher pool, indicating that the selection mechanism does not collapse to a single dominant teacher. We also observe that, under the same truncation strategy, a higher reference-exposure ratio does not necessarily produce a larger student-teacher KL distance, as discussed in \Cref{last}. This observation further suggests that existing adaptive methods of adjusting reference exposure alone cannot determine teacher suitability.


\textbf{Effect of Majority Voting and KL-Based Selection. } MTA-OPSD selects a teacher in two steps: majority voting and KL-based selection. \Cref{tab8} presents the contribution of each step, where the second row is our default setting. MTA-OPSD consistently outperforms all variants, further validating the effectiveness of our design: majority voting first identifies a reliable teacher subset, and KL-based selection chooses a teacher from that subset.

\section{Conclusion}
We present MTA-OPSD, the first token-level adaptive OPSD method. A teacher-intervention study shows that teacher preferences vary across tokens and motivates our two-step token-level adaptive teacher selection strategy: majority voting with KL-based selection. Future work may explore self-evolving teacher pools and more stable multi-teacher training.

\section*{Acknowledgments}
This paper is supported by the DAAD programme Konrad Zuse Schools of Excellence in Artificial Intelligence, sponsored by the Federal Ministry of Research, Technology and Space.

Qi Chen is supported by research funding from the ELLIS Institute Finland and the Department of Computer Science at Aalto University. 

We acknowledge CSC-IT Center for Science, Finland, for providing access to the supercomputer LUMI, owned by the European High Performance Computing Joint Undertaking (EuroHPC JU) and hosted by CSC Finland in collaboration with the LUMI consortium. We also acknowledge the computational resources provided by the Aalto Science-IT project through the Triton cluster.

\bibliography{ICLR2027}
\bibliographystyle{iclr2027_conference}

\newpage
\appendix
\section{Visualization Details} \label{a1}
\begin{figure}[h] 
    \centering 
    \includegraphics[width=1.0\textwidth]{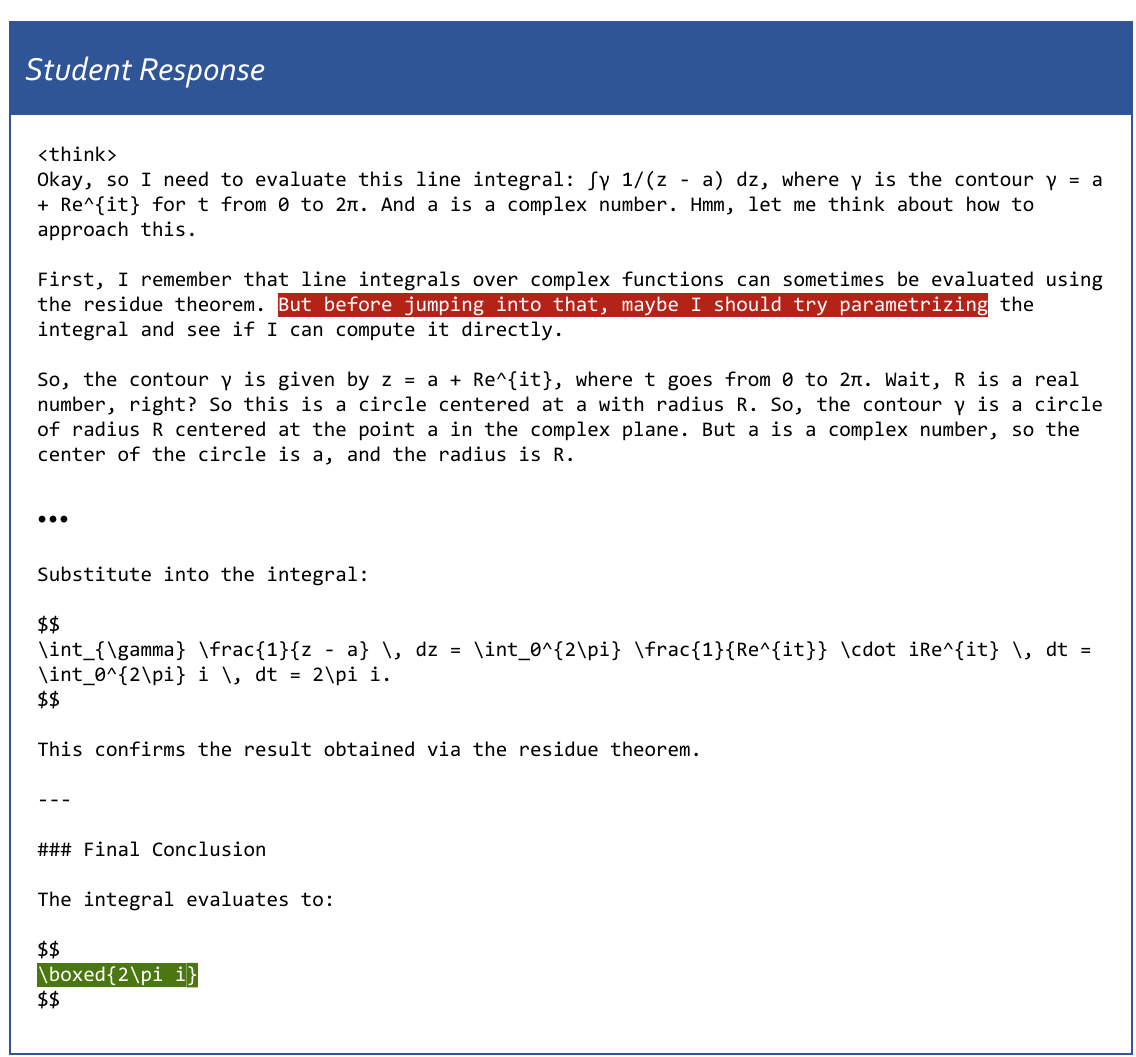}
    \vspace{-12pt}
    \caption{Entire student response for \Cref{fig1}.}
    \vspace{-0.2cm}
    \label{fig4}
\end{figure}

The three teachers in \Cref{fig1} use the configurations of prefix-$0.5$, suffix-$0.5$, and random-span-$0.5$. The entire student response is shown in \Cref{fig4}.

\section{Additional Majority Voting Details} \label{a2}
As discussed in \Cref{sec5.2}, a tie may occur when the multi-teacher pool contains an even number of teachers, with equal numbers of positive and negative votes. In this case, we break the tie according to the magnitudes of the corresponding advantage signals. Specifically, the majority direction is defined as
\begin{equation}
V_M(c_i)=
\begin{cases}
\operatorname{sgn} \left(
\displaystyle\sum_{k=1}^{K} v_{T_k}(c_i)
\right), \text{ if } \displaystyle\sum_{k=1}^{K} v_{T_k}(c_i)\neq 0,
\\
\operatorname{sgn}\left(
\displaystyle\sum_{k=1}^{K}
a_{T_k}(y^i\mid c_i)\right)
, \text{ otherwise}.
\end{cases}
\label{eq:majority_tie}
\end{equation}
Thus, when the numbers of positive and negative votes are equal, we select the group with the larger total absolute advantage. The majority teacher subset is then still constructed using \Cref{eq14}. This tie-breaking rule retains the direction supported by the stronger aggregate teacher signal.

\section{Additional Experiment Settings} \label{a3}

All experiments are conducted on 4 H200 GPUs. We use a learning rate of $1$ × $10^{-6}$. All methods are implemented on top of the veRL \citep{sheng2025hybridflow} PPO trainer with FSDP2 actor
sharding and vLLM rollouts. All runs use full-parameter training (no LoRA \citep{hu2021lora}). The global batch size is $32$. Following \citep{zhao2026self}, we restrict the token-level distillation objective to the top-100 entries of the teacher distribution.

Note that following \citep{pan2026rlcsd}, we use training problems at difficulty levels $5$–$7$ for Qwen3-1.7B, $6$–$8$ for Qwen3-4B, and levels $7$–$10$ for Qwen3-8B.

\section{Quality Visualization} \label{a4}
\begin{figure}[h] 
    \centering 
    \includegraphics[width=0.8\textwidth]{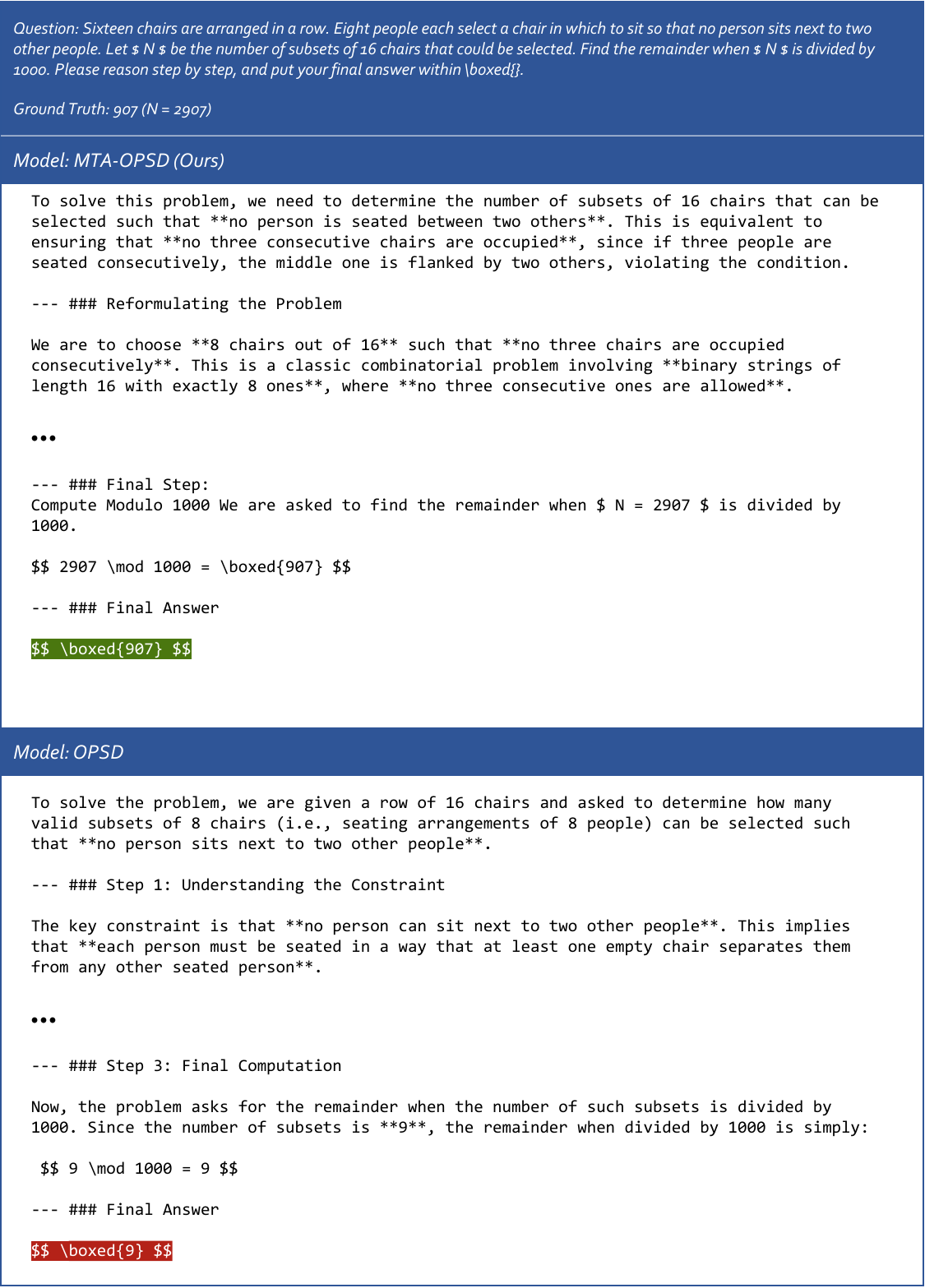}
    \vspace{-12pt}
    \caption{Quality visualization.}
    \vspace{-0.2cm}
    \label{fig5}
\end{figure}

We provide a qualitative comparison in \Cref{fig5}.

\section{Additional Details about KL-distance} \label{last}

As discussed in \Cref{sec6.3}, even under same truncation strategy, larger reference-exposure ratio does not lead to larger student-teacher KL distance. In our pilot study \Cref{sec4.2}, for example, for Qwen3-4B model with suffix truncation, only $49.1\%$ cases that $0.75$ exposure ratio has larger KL distance from the student. In contrast, the statistic for $0.5$ exposure ratio is $50.9\%$.

\section{Multi-teacher Pool Visualization} \label{a5}

Figures below demonstrate a brief visualization of our multi-teacher pool.
\begin{figure}[h] 
    \centering 
    \includegraphics[width=0.8\textwidth]{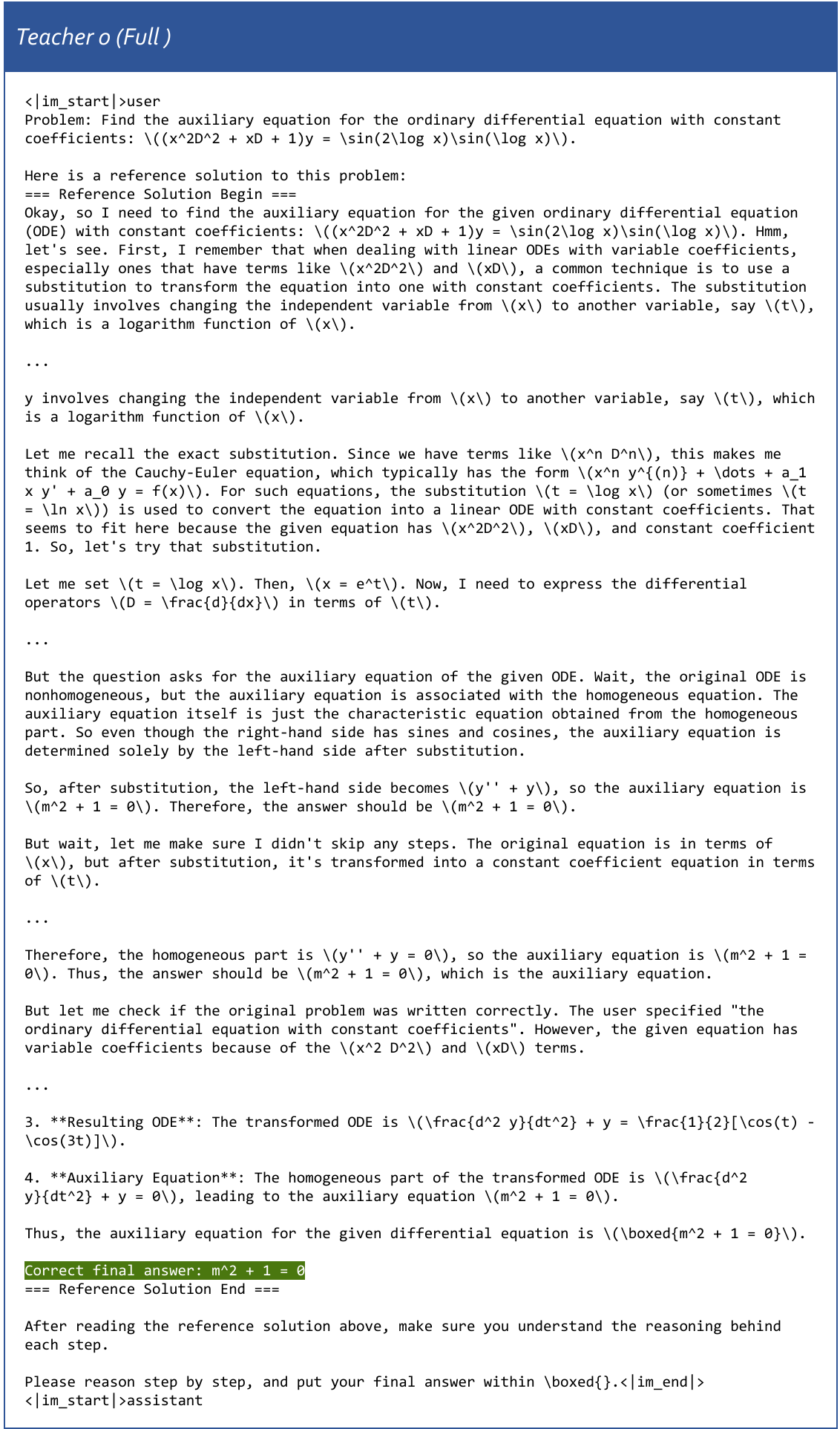}
    \vspace{-12pt}
    \caption{Multi-teacher pool visualization.}
    \vspace{-0.2cm}
    \label{fig6}
\end{figure}
\begin{figure}[h] 
    \centering 
    \includegraphics[width=0.8\textwidth]{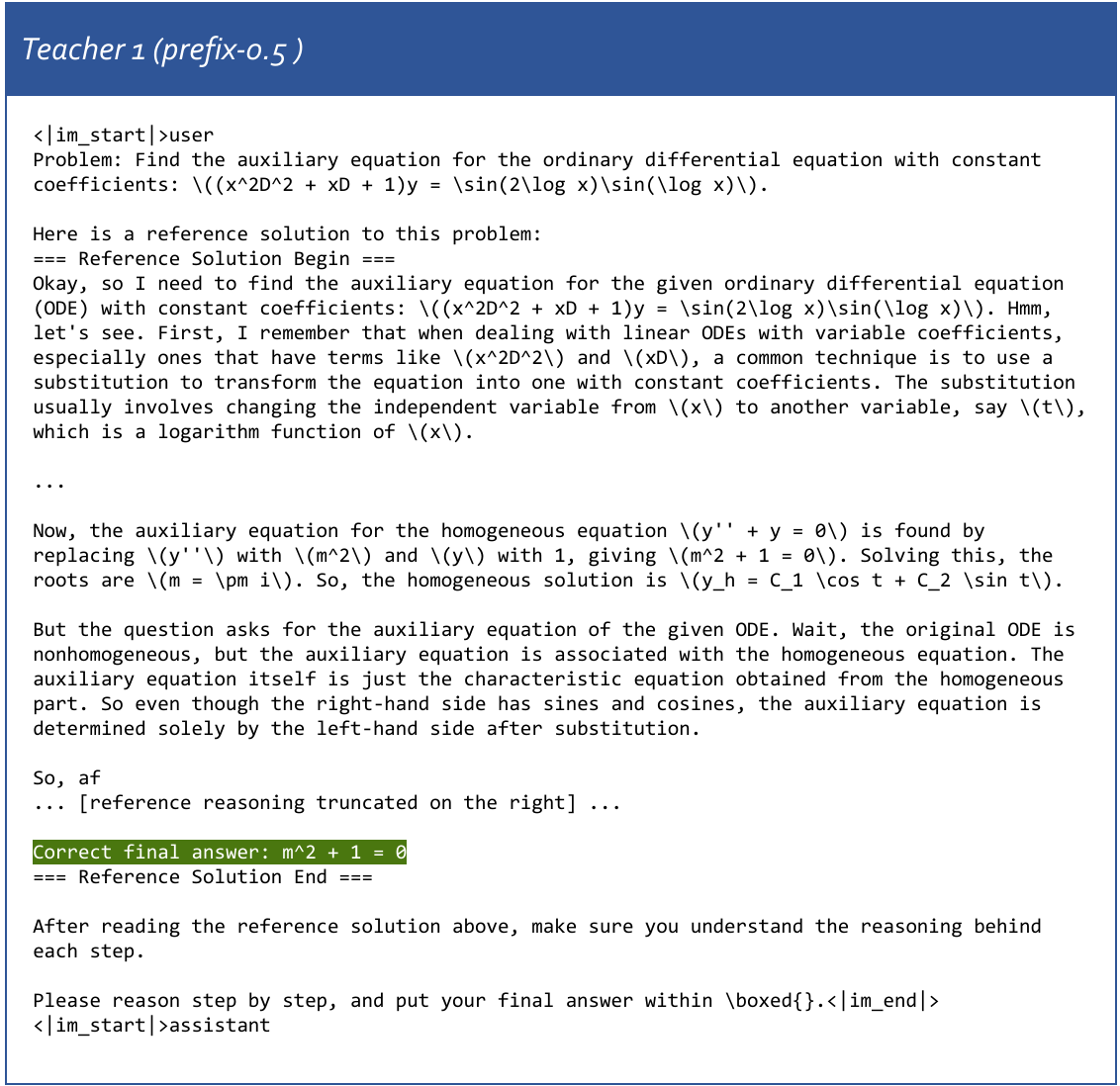}
    \vspace{-12pt}
    \caption{Multi-teacher pool visualization.}
    \vspace{-0.2cm}
    \label{fig7}
\end{figure}
\begin{figure}[h] 
    \centering 
    \includegraphics[width=0.8\textwidth]{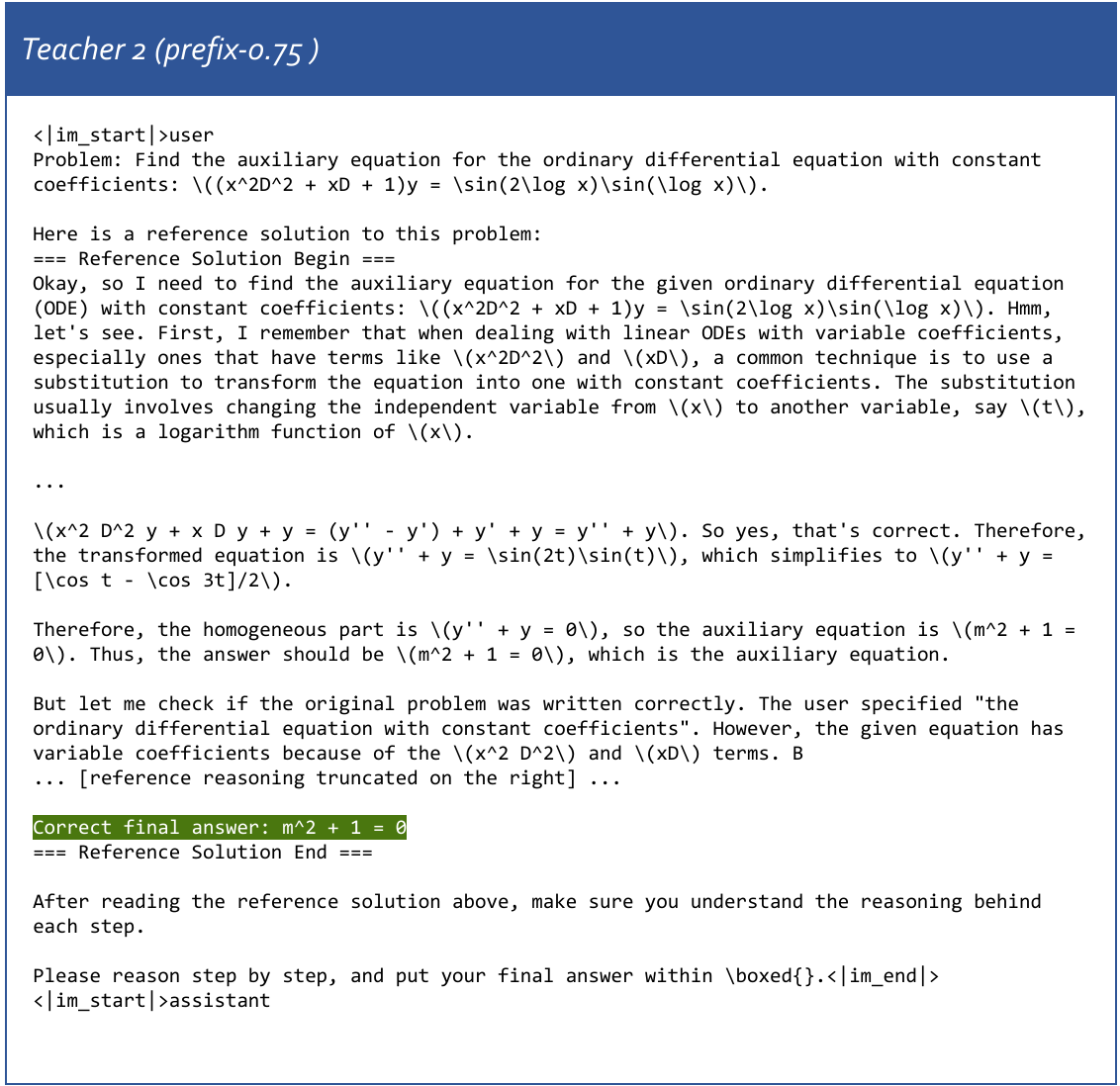}
    \vspace{-12pt}
    \caption{Multi-teacher pool visualization.}
    \vspace{-0.2cm}
    \label{fig8}
\end{figure}
\begin{figure}[h] 
    \centering 
    \includegraphics[width=0.8\textwidth]{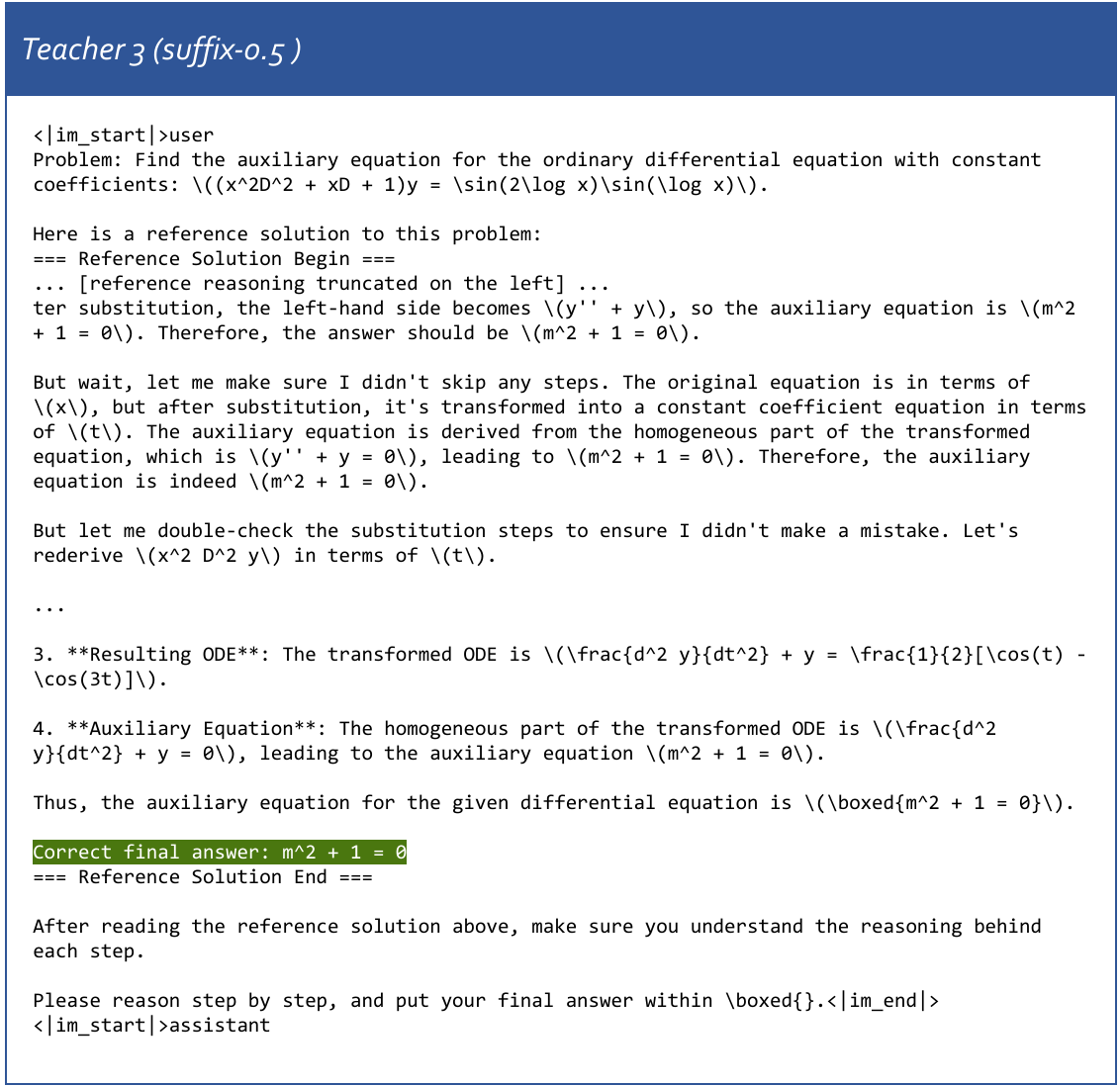}
    \vspace{-12pt}
    \caption{Multi-teacher pool visualization.}
    \vspace{-0.2cm}
    \label{fig9}
\end{figure}

\end{document}